%% file: emnlp2020.tex
\documentclass[11pt,a4paper]{article}
\usepackage[hyperref]{emnlp2020}
\usepackage{times}
\usepackage{latexsym}

\usepackage{amssymb}  

\usepackage{microtype}

\aclfinalcopy 
\def\aclpaperid{2462}

\usepackage{soul}
\usepackage{url}
\usepackage{graphicx}
\usepackage{amsmath}
\usepackage{amsthm}
\usepackage{booktabs}
\usepackage{algorithm}
\usepackage{algorithmic}
\usepackage{xcolor}

\definecolor{Red}{rgb}{1,0,0}
\definecolor{Green}{rgb}{0,0.7,0}
\definecolor{Blue}{rgb}{0,0,1}
\definecolor{Red}{rgb}{0.6,0,0}
\definecolor{Orange}{rgb}{1,0.5,0}

\definecolor{Red}{rgb}{1,0,0}
\newcommand{\camera}[1]{\textcolor{Red}{#1}}

\title{What-if I ask you to explain:\\ Explaining the effects of perturbations in procedural text}

\author{
  Dheeraj Rajagopal\textsuperscript{1}, Niket Tandon\textsuperscript{2}, Bhavana Dalvi\textsuperscript{2}, Peter Clark\textsuperscript{2}, Eduard Hovy\textsuperscript{1}\\
  \vspace{1mm} \\
        \textsuperscript{1}Carnegie Mellon University \hspace*{2mm}  \textsuperscript{2}Allen Institute for Artificial Intelligence   \\
        {\tt \{dheeraj,hovy\}@cs.cmu.edu \{niket,bhavanad,peterc\}@allenai.org}
}
\date{}
\begin{document}
\input{custom_commands.tex}
\maketitle

\begin{abstract}

Our goal is to explain the effects of perturbations in procedural text, e.g., given a passage describing a rabbit's life cycle, explain why illness (the perturbation) may reduce the rabbit population (the effect). Although modern systems are able to solve the original {\it prediction} task well (e.g., illness results in less rabbits), the {\it explanation} task - identifying the causal chain of events from perturbation to effect - remains largely unaddressed, and is the goal of this research.
%
%
We present \ourmodel~, a system that constructs such explanations from paragraphs, by modeling the explanation task as a multitask learning problem. \ourmodel~ constructs explanations from the sentences in the procedural text, achieving $\sim$ 18 points better on explanation accuracy compared to several strong baselines on a recent process comprehension benchmark. On an end task on this benchmark, we show a surprising finding that good explanations do not have to come at the expense of end task performance, in fact leading to a 7\% F1 improvement over SOTA.
\end{abstract}

\input{sections/introduction.tex}

\input{sections/related_work.tex}
\input{sections/problem_formulation.tex}
\input{sections/quartet_model.tex}
\input{sections/experiments.tex}

\input{sections/downstream_qa.tex}

\input{sections/error_analysis.tex}
\input{sections/conclusion.tex}

\section*{Acknowledgements}

We thank Harsh Jhamtani, Keisuke Sakaguchi, Vidhisha Balachandran, Dongyeop Kang,  members of the AI2 Aristo group, and the anonymous reviewers for their
insightful feedback.

\bibliography{emnlp2020}
\bibliographystyle{acl_natbib}

\input{sections/appendix}

\end{document}

%% file: custom_commands.tex
\newcommand{\ourmodel}[1]{\textsc{Quartet}}
\newcommand{\taggingmodel}[1]{\textsc{Tagging}}
\newcommand{\ourmodelplus}[1]{\textsc{Quartet+}}
\newcommand{\stepwisemodel}[1]{\textsc{Stepwise}}
\newcommand{\wiqabert}[1]{\textsc{bert-no-expl}}
\newcommand{\ourdata}[1]{\textsc{wiqa}}
\newcommand{\ourmodelplusnoexpl}[1]{\textsc{bert-w/-expl}}
\newcommand{\effectonly}[1]{\textsc{$q_e$only}}
\newcommand{\DataAug}[1]{\textsc{DataAug}}

\definecolor{Red}{rgb}{1,0,0}
\definecolor{Green}{rgb}{0,0.7,0}
\definecolor{Blue}{rgb}{0,0,1}
\definecolor{Red}{rgb}{0.6,0,0}
\definecolor{Orange}{rgb}{1,0.5,0}
\newcommand{\reviewed}[1]{{#1}}

\makeatletter
\newcommand*\bigcdot{\mathpalette\bigcdot@{.5}}
\newcommand*\bigcdot@[2]{\mathbin{\vcenter{\hbox{\scalebox{#2}{$\m@th#1\bullet$}}}}}
\makeatother

\renewcommand{\v}[1]{$\mathbf{#1}$}
\newcommand{\vect}[1]{\mathbf{#1}}
\newcommand{\cor}[1]{${\mathbf{+}}$}
\newcommand{\pos}[1]{${\mathbf{+}}$}
\newcommand{\opp}[1]{${\mathbf{-}}$}
\newcommand{\no}[1]{${\bigcdot}$}

\newcommand{\bluebox}[1]{\colorbox{blue!10}{#1}}
\newcommand{\redbox}[1]{\colorbox{red!10}{#1}}
\newcommand{\purplebox}[1]{\colorbox{purple!10}{#1}}



\def\DG{{\mathcal{G}}}

\newtheorem{theorem}{Definition}[section]

\newcommand{\statechange}[1]{\texttt{\textit{#1}}}
\newcommand{\entity}[1]{\texttt{#1}}
\newcommand{\strikethrough}[1]{\st{#1}}

\newcommand{\namecite}[1]{\citeauthor{#1}~\shortcite{#1}}
\newcommand{\com}[1]{}
\newcommand{\myparagraph}[1]{\vspace{1mm} \noindent {\bf #1: }}
\newcommand{\bfit}[1]{\textbf{\textit{#1}}}
\newcommand{\eat}[1]{}
\mathchardef\mhyphen="2D
\newenvironment{ite}{                     
     \parskip 0cm \begin{itemize} \parskip 0cm \parsep 0cm \itemsep 0cm \topsep 0cm}{
        \end{itemize}} 
\newenvironment{enu}{                   
     \parskip 0cm \begin{list}{}{\parsep 0cm \itemsep 0cm \topsep 0cm}}{
      \end{list}} 
\newenvironment{des}{                 
     \parskip 0cm \begin{list}{}{\parsep 0cm \itemsep 0cm \topsep 0cm}}{
      \end{list}} 
\newenvironment{myenumerate}{                   
     \parskip 0cm \begin{enumerate}{\parsep 0cm \itemsep 0cm \topsep 0cm}}{
        \end{enumerate}} 
\newenvironment{myitemize}{                     
     \parskip 0cm \begin{itemize}{\parsep 0cm \itemsep 0cm \topsep 0cm}}{
        \end{itemize}} 
\newcommand{\ourdataexpansion}{``What-If Question Answering''}
\newenvironment{myquote}{                   
  \parskip 0mm \begin{quoting}[vskip=0mm,leftmargin=2mm]}{
\end{quoting}}
\newcommand{\red}[1]{\textcolor{red}{#1}}
\newcommand{\blue}[1]{\textcolor{blue}{#1}}
\newcommand{\green}[1]{\textcolor{green}{#1}}
\newenvironment{mycentering}
 {\parskip=0pt\par\nopagebreak\centering}
 {\par\noindent\ignorespacesafterend}

    

\newcommand{\squishlist}{
  \begin{list}{$\bullet$}
    { \setlength{\itemsep}{0pt}      \setlength{\parsep}{3pt}
      \setlength{\topsep}{3pt}       \setlength{\partopsep}{0pt}
      \setlength{\leftmargin}{1.5em} \setlength{\labelwidth}{1em}
      \setlength{\labelsep}{0.5em} } }
\newcommand{\reallysquishlist}{
  \begin{list}{$\bullet$}
    { \setlength{\itemsep}{0pt}    \setlength{\parsep}{0pt}
      \setlength{\topsep}{0pt}     \setlength{\partopsep}{0pt}
      \setlength{\leftmargin}{0.2em} \setlength{\labelwidth}{0.2em}
      \setlength{\labelsep}{0.2em} } }

 \newcommand{\squishend}{
     \end{list} 
 }
 
 \newcommand{\squishenu}{
  \begin{enumerate}
    { \setlength{\itemsep}{0pt}    \setlength{\parsep}{0pt}
      \setlength{\topsep}{0pt}     \setlength{\partopsep}{0pt}
      \setlength{\leftmargin}{0.2em} \setlength{\labelwidth}{0.2em}
      \setlength{\labelsep}{0.2em} } }

 \newcommand{\squishenuend}{
     \end{enumerate} 
 }

%% file: sections/introduction.tex
\section{Introduction}
\label{sec:introduction}
Procedural text is common in natural language (in recipes, how-to guides, etc.)
and finds many applications such as automatic execution of biology experiments \cite{mysore2019materials}, cooking recipes \cite{Bollini2012cookingrobot} and everyday activities \cite{Wikihow-sigir2015}. However, the goal of procedural text understanding in these settings remains a major challenge and requires two key abilities, (i) understanding the dynamics of the world \textit{inside} a procedure by tracking entities and what events happen as the narrative unfolds. (ii) understanding the dynamics of the world \textit{outside} the procedure that can influence the procedure.

While recent systems for procedural text comprehension have focused on understanding the dynamics of the world \textit{inside} the process, such as tracking entities and answering questions about what events happen, e.g., \cite{propara-emnlp18,npn,Henaff2016TrackingTW}, the extent to which they understand the influences of \textit{outside} events remains unclear. In particular, if a system fully understands a process, it should be able to predict what would happen if it was perturbed in some way due to an event from the \textit{outside} world. Such counterfactual reasoning is particularly challenging because, rather than asking what happened (described in text), it asks about what \textbf{would} happen in an alternative world where the change occurred.

\begin{figure}[!t]
    {\includegraphics[width=1.05\columnwidth]{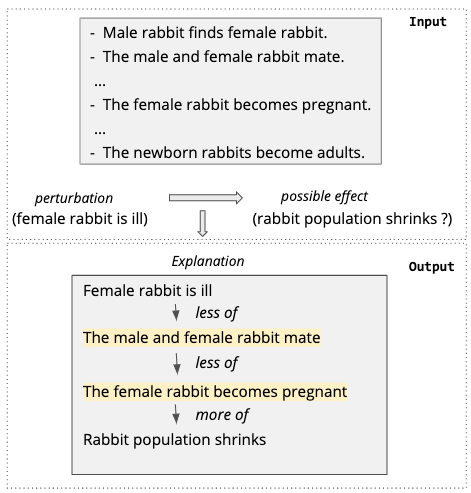}}
    \caption{Given a procedural text, the task is to explain the effect of the perturbation using the input sentences.
    }
    \label{fig:running-example}
\end{figure}


Recently, \citet{wiqa} introduced the WIQA dataset that contains such problems, requiring prediction of the effect of perturbations in a procedural text. They also presented several strong models on this task. However, it is unclear whether those high scores indicate that the models fully understand the described procedures, i.e., that the models have knowledge of the causal chain from perturbation to effect. To test this, \citet{wiqa} also proposed an explanation task. While the general problem of synthesizing explanations is hard, they proposed a simplified version in which explanations were instead assembled from sentences in the input paragraph and qualitative indicators (more/less/unchanged). Although they introduced this explanation task and dataset, they did not present a model to address it. We fill this gap by proposing the first solution to this task.

We present a model, \ourmodel~ (QUAlitative Reasoning wiTh ExplanaTions) that takes as input a passage and a perturbation, and its qualitative effect. The output contains the qualitative effect and an explanation structure over the passage. See Figure \ref{fig:running-example} for an example.  The explanation structure includes up to two supporting sentences from the procedural text, together with the qualitative effect of the perturbation on the supporting sentences (\emph{more of} or \emph{less of} in Figure \ref{fig:running-example}). \ourmodel~ models this qualitative reasoning task as a multitask learning problem to explain the effect of a perturbation. 

Our main contributions are:
\squishlist
    \item We present the first model that explains the effects of perturbations in procedural text. On a recent process comprehension benchmark, \ourmodel~ generates better explanations compared to strong baselines.
    
    \item On an end task on this benchmark, we show a finding that good explanations do not have to come at the expense of end task performance, in fact leading to a 7\% F1 improvement over SOTA. 
    (refer \S \ref{sec:downstream_qa}). Prior work has found that optimizing for explanation can hurt end-task performance. Ours is a useful datapoint showing that good explanations do not have to come at the expense of end-task performance\footnote{All the code will be publicly shared upon acceptance}.
    
\squishend

\begin{table*}[ht]
\centering
\resizebox{1.0\textwidth}{!}
{%
\begin{tabular}{|l|l|}
\hline
    
    ears less protected  $\rightarrow$ \blue{\texttt{(MORE/+)}} \bluebox{sound enters the ear} $\rightarrow$ \blue{\texttt{(MORE/+)}} \bluebox{sound hits ear drum} $\rightarrow$ \blue{\texttt{(MORE/+)}} more sound detected \\
    
    blood clotting disorder  $\rightarrow$ \blue{\texttt{(LESS/-)}} \bluebox{blood clots} $\rightarrow$ \blue{\texttt{(LESS/-)}} \bluebox{scab forms} $\rightarrow$ \blue{\texttt{(MORE/+)}} less scab formation \\
    
    breathing exercise  $\rightarrow$ \blue{\texttt{(MORE/+)}} \bluebox{air enters lungs} $\rightarrow$ \blue{\texttt{(MORE/+)}} \bluebox{air enters windpipe} $\rightarrow$ \blue{\texttt{(MORE/+)}} oxygen enters bloodstream \\
    
    squirrels store food  $\rightarrow$ \blue{\texttt{(MORE/+)}} \bluebox{squirrels eat more} $\rightarrow$ \blue{\texttt{(MORE/+)}} \bluebox{squirrels gain weight} $\rightarrow$ \blue{\texttt{(MORE/+)}} hard survival in winter \\

    less trucks run  $\rightarrow$ \blue{\texttt{(LESS/-)}} \bluebox{trucks go to refineries} $\rightarrow$ \blue{\texttt{(LESS/-)}} \bluebox{trucks carry oil} $\rightarrow$ \blue{\texttt{(MORE/+)}} less fuel in gas stations \\
    
    coal is expensive  $\rightarrow$ \blue{\texttt{(LESS/-)}} \bluebox{coal burns} $\rightarrow$ \blue{\texttt{(LESS/-)}} \bluebox{heat produced from coal} $\rightarrow$ \blue{\texttt{(LESS/-)}} electricity produced \\
    
    legible address  $\rightarrow$ \blue{\texttt{(MORE/+)}} \bluebox{mailman reads address} $\rightarrow$ \blue{\texttt{(MORE/+)}} \bluebox{mail reaches destination} $\rightarrow$ \blue{\texttt{(MORE/+)}} on-time delivery \\
    
    more water to roots  $\rightarrow$ \blue{\texttt{(MORE/+)}} \bluebox{root attract water} $\rightarrow$ \blue{\texttt{MORE/+)}} \bluebox{roots suck up water} $\rightarrow$ \blue{\texttt{(LESS/-)}} plants malnourished \\

    in a quiet place  $\rightarrow$ \blue{\texttt{(LESS/-)}} \bluebox{sound enters the ear} $\rightarrow$ \blue{\texttt{(LESS/-)}} \bluebox{sound hits ear drum} $\rightarrow$ \blue{\texttt{(LESS/-)}} more sound detected \\
    
    eagle hungry  $\rightarrow$ \blue{\texttt{(MORE/+)}} \bluebox{eagle swoops down} $\rightarrow$ \blue{\texttt{(MORE/+)}} \bluebox{eagle catches mouse} $\rightarrow$ \blue{\texttt{(MORE/+)}} eagle gets more food \\
    \hline
    \end{tabular}%
    }
\caption{Examples of our model's predictions on the dev. set in the format: ``$q_p \rightarrow $ \blue{$d_i$} \bluebox{$x_i$} $\rightarrow$ \blue{$d_j$} \bluebox{$x_j$} $\rightarrow$ \blue{$d_e$} $q_e$''. Supporting sentences $x_i$, $x_j$ are compressed e.g., ``the person has his ears less protected'' $\rightarrow$ ``ears less protected''}
\label{tab:many-examples}
\end{table*}

%% file: sections/related_work.tex
\section{Related work}
\label{sec:related-work}

\textbf{Procedural text understanding:} Machine reading has seen tremendous progress. With machines reaching human performance in standard QA benchmarks \cite{Devlin2018BERT,rajpurkar2016squad}, more challenging datasets have been proposed \cite{Dua2019DROP} that require background knowledge,
commonsense reasoning \cite{talmor2019commonsenseqa} and visual reasoning \cite{Antol2015VQA,Zellers2018VCR}. In the context of procedural text understanding which has gained considerable amount of attention recently,  \cite{npn,Henaff2016TrackingTW,propara-naacl18} address the task of tracking entity states throughout the text. Recently, \cite{wiqa} introduced the \textsc{wiqa} task to predict the effect of \textit{perturbations}.

Understanding the effects of perturbations, specifically, qualitative change, has been studied using formal frameworks in the qualitative reasoning community \cite{Forbus1984QualitativePT,weld2013readings} and counterfactual reasoning in the logic community \cite{lewis2013counterfactuals}. The \ourdata~ dataset situates this task in terms of natural language rather than formal reasoning, by treating the task as a mixture of reading comprehension and commonsense reasoning. However, existing models do not explain the effects of perturbations.

\noindent\textbf{Explanations:} Despite large-scale QA benchmarks, high scores do not necessarily reflect understanding \cite{min2019singlehotpot}. Current models may not be robust or exploit annotation artifacts \cite{gururangan2018annotationartifacts}. This makes explanations desirable for interpretation \cite{selvaraju2017gradcamExplanationsHelp}.

Attention based explanation has been successfully used in vision tasks such as object detection \cite{petsiuk2018RISEAttentionExplanations} because pixel information is explainable to humans. These and other token level attention models used in NLP tasks \cite{wiegreffe2019attentionisnotnotexplanation} do not provide full-sentence explanations of a model's decisions.

Recently, several datasets with natural language explanations have been introduced, e.g., in natural language inference \cite{Camburu2018eSNLI}, visual question answering \cite{Park2018MultimodalExplanationsDataset}, and multi-hop reading comprehension (HotpotQA dataset) \cite{Yang2018HotpotQA}. In contrast to these datasets, we explain the effects of perturbations in procedural text. HotpotQA contains explanations based on two sentences from a Wikipedia paragraph. Models on the HotpotQA would not be directly applicable to our task and require substantial modification for the following reasons: (i) HotpotQA models are not trained to predict the qualitative structure (more or less of chosen explanation sentences in Figure \ref{fig:running-example}). (ii) HotpotQA involves reasoning over named entities, whereas the current task focuses on common nouns and actions (models that work well on named entities need to be adapted to common nouns and actions \cite{hanie-common-nouns}). (iii) explanation paragraphs in HotpotQA are not procedural while the current input is procedural in nature with a specific chronological structure.

Another line of work provides more structure and organization to explanations, e.g., using scene graphs in computer vision \cite{Ghosh2019NaturalLangExplanationsVQA}. For elementary science questions, \citet{jansen2018worldtree} uses a science knowledge graph. These approaches rely on a knowledge structure or graph but knowledge graphs are incomplete and costly to construct for every domain \cite{weikum2010pods}. There are trade-offs between unstructured and structured explanations. Unstructured explanations are available abundantly while structured explanations need to be constructed and hence are less scalable \cite{Camburu2018eSNLI}. Generating free-form (unstructured) explanations is difficult to evaluate \cite{Cui2018LearningToEvaluate,Zhang2019BERTScoreET}, and adding qualitative structure over them is non-trivial. Taking a middle ground between free-form and knowledge graphs based explanations, we infer a qualitative structure over the sentences in the paragraph. This retains the rich interpretability and simpler evaluation of structured explanations as well as leverages the large-scale availability of sentences required for these explanation.

It is an open research problem whether requiring explanation helps or hurts the original task being explained. On the natural language inference task (e-SNLI), \citet{Camburu2018eSNLI} observed that models generate correct explanations \textit{at the expense of} good performance. On the Cos-E task, recently  \citet{rajani2019salesforceexplain} showed that explanations help the end-task. Our work extends along this line in a new task setting that involves perturbations and enriches natural language explanations with qualitative structure.

%% file: sections/problem_formulation.tex
\section{Problem definition}
\label{sec:problemdef}

We adopt the problem definition described in \citet{wiqa}, and summarize it here.

\paragraph{Input:} 1. Procedural text with steps $x_1 \dots x_K$. Here, $x_k$ denotes step $k$ (i.e., a sentence) in a procedural text comprising $K$ steps. \\
2. A perturbation $q_p$ to the procedural text and its likely candidate effect $q_e$.

\paragraph{Output:}

An explanation structure that explains the effect  of the perturbation $q_p$:
    $$q_{p} \rightarrow d_i x_i \rightarrow d_j x_j \rightarrow d_e q_e$$
    
    \squishlist
        \item $i$: step id for the first supporting sentence. 
        \item $j$: step id for the second supporting sentence.
        \item $d_i$ $\in$ \{\cor, \opp, \no~ \}: how step id $i$ is affected.
        \item $d_j$ $\in$ \{\cor, \opp, \no~ \}: how step id $j$ is affected.
        \item $d_e$ $\in$ \{\cor, \opp, \no~ \}: how $q_e$ is affected.
    \squishend
    
See Figure \ref{fig:running-example} for an example of the task, and Table \ref{tab:many-examples} for examples of explanations.
    
An explanation consists of up to two (i.e., zero, one or two) supporting sentences $i, j$ along with their qualitative directions $d_i ,d_j$. If there is only one supporting sentence, then $j$ = $i$. If $d_e= $ \no~  , then $i=$\O, $j=$\O~ (there is no valid explanation for no-effect). 
    
While there can be potentially many correct explanation paths in a passage, the \ourdata~ dataset consists of only one gold explanation considered best by human annotators. Our task is to predict that particular gold explanation.    
    
\paragraph{Assumptions:} In a procedural text, steps $x_1 \dots x_K$ are chronologically ordered and have a forward flowing effect i.e., if $j > i$ then more/increase of $x_i$ will result in more/increase of $x_j$. Prior work on procedural text makes a similar assumption \cite{propara-naacl18}. Note that this assumption does not hold for cyclic processes, and cyclic processes have already been flattened in \ourdata~ dataset. We make the following observations based on this \textit{forward-flow} assumption.
    \squishlist
        \item[a1:] $i <= j$ (\textit{forward-flow} order)
        \item[a2:] $d_j = d_i$ (\textit{forward-flow} assumption)\camera{
        \footnote{Note that this does not assume all sentences have the same directionality of influence. For example, a paragraph could include both positive and negative influences: ``Predators arrive. Thus the rabbit population falls...''. Rather, the $d_j = d_i$ assumption is one of narrative coherence: the {\it more} predators arrive, the {\it more} the rabbit population falls. That is, within a paragraph, we assume enhancing one step will have enhanced effects (both positive or negative effects) on future steps - a property of a coherently authored paragraph.}}
        \item[a3:] For the WIQA task, $d_e$ is the answer label because it is the end node in the explanation structure.
        \item[a4:] If $d_i=$ \no~ 
        then answer label $=$ \no~ (since $q_p$ does not affect $q_e$, there is no valid explanation.) 
        \item[a5:] $1 \leq i \leq K$; if $d_i = $ \no~, then i =  \O~ (see a4) 
        \item[a6:] $i \leq j \leq K$; if $d_e = $ \no~, then j =  \O~ (see a4)
    \squishend

This assumption reduces the number of predictions, removing $d_j$ and answer label (see a2, a3). Given $x_1 \dots x_K$, $q_p$, $q_e$ the model must predict four labels: $i$, $j$, $d_i$, $d_e$ . 

%% file: sections/quartet_model.tex
\section{\ourmodel~ model}
\label{sec:model-quartet}

We can solve the problem as a classification task, predicting four labels: $i$, $j$, $d_i$, $d_e$. If these predictions are performed independently, it requires several independent classifications and this can cause error propagation: prediction errors that are made in the initial stages cannot be fixed and can propagate into larger errors later on \cite{goldberg2017neural}. 

\begin{figure*}[!h]
    {\includegraphics[width=0.95\textwidth]{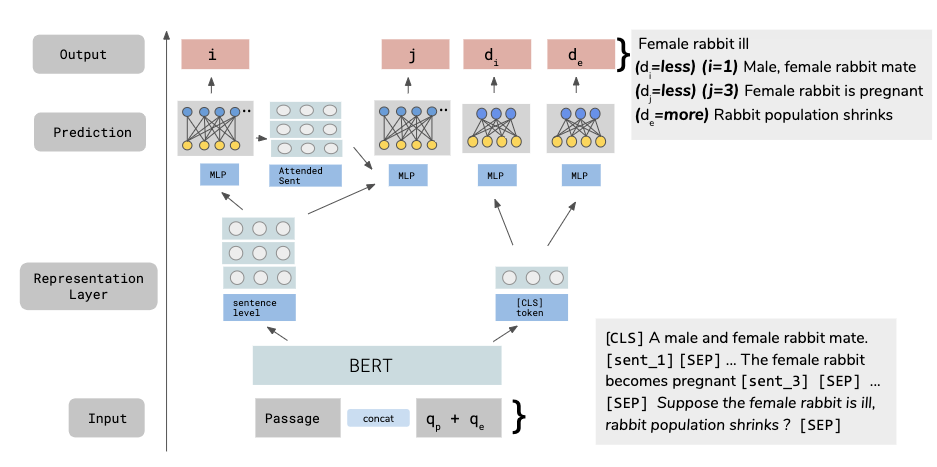}}
    \caption{\ourmodel~ model. \textit{Input}: Concatenated passage and question using standard BERT word-piece tokenization. \textit{Representation Layer}: The input is encoded using BERT transformer. We obtain \texttt{[CLS]} and sentence level representations. \textit{Prediction}: From the sentence level representation, we use an MLP to model the distributions for $i$ and $j$ (using attended sentence representation). From \texttt{[CLS]} representation, we use MLP for $d_i$ (and $d_j$, since $d_i=d_j$) and $d_e$ distributions. \textit{Output}: Softmax to predict $\{ i, j, d_i, d_j, d_e \}$ \ 
    }
    \label{fig:model}
\end{figure*}

To avoid this, \ourmodel~ predicts and explains the effect of $q_p$ as a multitask learning problem, where the representation layer is shared across different tasks. We apply the widely used parameter sharing approach, where a single representation layer is followed by task specific output layers \cite{Baxter1997}. This reduces the risk of overfitting to a single task and allows decisions on $i, j, d_i, d_e$ to influence each other in the hidden layers of the network. We first describe our encoder and then the other layers on top, see Figure \ref{fig:model} for the model architecture. 

\paragraph{Encoder:} To encode $x_1 \dots x_K$ and question $q$ we use the \textsc{BERT}
architecture \cite{Devlin2018BERT} that has achieved state-of-the-art performance across several NLP tasks \cite{clark2019bertanalysis}, where the question $q = q_p \oplus q_e$ ($\oplus$ stands for concatenation). We start with a byte-pair tokenization \cite{sennrich2015neural} of the concatenated passage and question ($x_1 \dots x_K \oplus q$) .  Let $[x_k]$ denote the byte-pair tokens of sentence $x_k$. The text is encoded as
\texttt{[CLS] $[x_1]$ $[unused1]$ [SEP] $[x_2]$ $[unused2]$ [SEP] .. $[q]$ [SEP]}. Here, \texttt{[CLS]} indicates a special classification token. \texttt{[SEP]} and $[unused1..K]$ are special next sentence prediction tokens.

These byte-pair tokens are passed through a 12-layered Transformer network, resulting in a contextualized representation for every byte-pair token. In this contextualized representation, the vector $\vect{u} = [\vect{u_1}, ... \vect{u_K}, \vect{u_q}] $ where $\vect{u_k}$ denotes the encoding for $[x_k]$, and $\vect{u_q}$ denotes question encoding. Let $E^l$ be the embedding size resulting from $l^{th}$ transformer layer. In that $l^{th}$ layer, $[\vect{u_1}, ... \vect{u_K}] \in \mathbb{R}^{K*E^l}$. The hidden representation of all transformer layers are initialized with weights from a self-supervised pre-training phase, in line with contemporary research that uses pre-trained language models \cite{Devlin2018BERT}.

To compute the final logits, we add a linear layer over the different transformer layers in BERT that are individual winners for individual tasks in our multitask problem. For instance, out of the total 12 transformer layers, lower layers (layer 2) are the best predictors for $[i, j]$ while upper layers (layer 10 and 11) are the best performing predictors for $[d_i, d_e]$. \citet{Zhang2019BERTScoreET} found that the last layer is not necessarily the best performing layer. Different layers seem to learn complementary information because their fusion helps. Combining different layers by weighted averaging of the layers has been attempted with mixed success \cite{Zhang2019BERTScoreET,clark2019bertanalysis}. We observed the same trend for simple weighted transformation. However, we found that learning a linear layer over concatenated features from winning layers improves performance. This is probably because there is very different information encoded in a particular dimension across different layers, and the concatenation preserves it better than simple weighted averaging.

\paragraph{Classification tasks:}
To predict the first supporting sentence $x_i$, we obtain a softmax distribution $s_i \in \mathbb{R}^{K}$ over $[\vect{u_1}, ... \vect{u_K}]$. From the \textit{forward-flow} assumption made in the problem definition section earlier, we know that $i \leq j$, making it possible to model this as a span prediction $x_{i:j}$. Inline with standard span based prediction models  \cite{Seo2016BidirectionalAF}, we use an attended sentence representation  $(s_i \odot [\vect{u_1}, ... \vect{u_K}]) \oplus  ([\vect{u_1}, ... \vect{u_K}]) \in \mathbb{R}^{K*2E^l}$ to predict a softmax distribution $s_j \in \mathbb{R}^{K}$ to obtain $x_j$. Here, $\odot$ denotes element-wise multiplication and $\oplus$ denotes concatenation.

For classification of $d_i$ (and $d_j$, since $d_i = d_j$), we use the representation of the first token (i.e., \texttt{CLS} token $ \in \mathbb{R}^{E^l}$) and a linear layer followed by softmax to predict $d_i \in$  \{ \cor, \opp, \no~ \}. Classification of $d_e$ is performed in exactly the same manner.

The network is trained end-to-end to minimize the sum of cross-entropy losses for the individual classification tasks $i, j, d_i, d_e$. At prediction time, we leverage assumptions (a4, a5, a6) to generate consistent predictions.

%% file: sections/experiments.tex
\section{Experiments}
\label{sec:experiments}

\paragraph{Dataset:} We train and evaluate \ourmodel~ on the recently published \ourdata~ dataset \footnote{\small{WIQA dataset link: \url{http://data.allenai.org/wiqa/}}} comprising of 30,099 questions from 2107 paragraphs with explanations (23K train, 5K dev, 2.5K test). The perturbations $q_p$ are either linguistic variation (17\% examples) of a passage sentence (these are called in-para questions) or require commonsense reasoning to connect to a passage sentence (41\% examples) (called, out-of-para questions). 
Explanations are supported by up to two sentences from the passage: 52.7\% length 2, 5.5\% length 1, 41.8\% length 0. Length zero explanations indicate that $d_e = $\no~ (called, no-effect questions), and ensure that random guessing on explanations gets low score on the end task. 
    
\paragraph{Metrics:} We evaluate on both explainability and the downstream end task (QA). For explainability, we define explanation accuracy as the average accuracy of the four components of the explanation: $acc_{expl} = \frac{1}{4} * \sum_{i \in \{i,j, d_i, d_e \}} acc(i)$ and $acc_{qa} = acc(d_e)$ (by assumption a3). The QA task is measured in terms of accuracy. 

\paragraph{Hyperparameters:} \ourmodel~ fine-tunes  \textsc{BERT}, allowing us to re-use the same hyperparameters as \textsc{BERT} with small adjustments in the recommended range \cite{Devlin2018BERT}. We use the \textsc{BERT}-base-uncased version with a hidden size of 768. We use the standard adam optimizer with a learning rate 1e-05, weight decay 0.01, and dropout 0.2 across all the layers\footnote{Hyperparameter search details in appendix \S \ref{subsec:appendix_hyperparam}}. 
All the models are trained on an NVIDIA V-100 GPU.

\paragraph{Models:} We measure the performance of the following baselines (two non-neural and three neural).

    \reallysquishlist
        \item \textsc{Random}: Randomly predicts one of the three labels \{\cor, \opp, \no~ \} to guess $[d_i, d_e]$. Supporting sentences $i$ and $j$ are picked randomly from  $|avg_{sent}|$ sentences.
        
        \item \textsc{Majority}: Predicts the most frequent label ({\it no effect i.e. $d_e$=\no~} in the case of \ourdata~ dataset.) 

        \item \effectonly~  : Inspired by existing works \cite{gururangan2018annotationartifacts}, this baseline exploits annotation artifacts (if any) in the explanation dataset by retraining \ourmodel~ using only $q_e$ while hiding the permutation $q_p$ in the question.

        \item 
        \textsc{Human} upper bound (Krippendorff's alpha  inter-annotator values on $[i, j, d_i]$) on explainability reported in \cite{wiqa}\footnote{https://allenai.org/data/wiqa}.

        \item \taggingmodel~: We can reduce our task to a structured prediction task. An explanation $i, j, d_i, d_e$ requires span prediction $x_{i:j}$ and labels on that span. So, for example, the explanation $i=1, j=2, d_i = $\cor~$, d_j =$\opp~ for input $x_1 \cdot x_5$  can be expressed as a tag sequence: \textsc{\texttt{B-CORRECT E-OPPOSITE O O O}}. Explanation $i=2, j=4, d_i = $\cor~, $d_j =$\opp~  would be expressed as: \textsc{\texttt{O B-CORRECT I-CORRECT E-OPPOSITE O}}. When $d_e$ = \no~ , then the tag sequence will \texttt{O O O O O}. This BIEO tagging scheme has seven labels $T$ = \{\texttt{\textsc{B-CORRECT, I-CORRECT, B-OPPOSITE, I-OPPOSITE, E-CORRECT, E-OPPOSITE, O}}\}.
        
        Formulating as a sequence tagging task allows us to use any standard sequence tagging model such as CRF as baseline. The decoder invalidates sequences that violate assumptions (a3 - a6). To make the encoder strong and yet comparable to our model, we use exactly the same \textsc{BERT} encoder as \ourmodel~. 
        For each sentence representation $u_k$, we predict a tag $\in T$. A CRF over these local predictions  additionally provides global consistency. The model is trained end-to-end by minimizing the negative log likelihood from the CRF layer.

        \item \wiqabert~: State-of-the-art BERT model \cite{wiqa} that only predicts the final answer $d_e$, but cannot predict the explanation. 
        
        \item \ourmodelplusnoexpl~: A standard BERT based approach to the explanation task that predicts the explanation structure. This model minimizes only the cross-entropy loss of the final answer $d_e$, predicting an explanation that provides the best answer accuracy.
        
        \item \DataAug~: This baseline is adapted from \citet{asai-hajishirzi-2020-logic}, where a RoBERTa model is augmented with symbolic knowledge and uses an additional consistency-based regularizer. Compared to our model, this approach uses a more robustly pre-trained BERT (RoBERTa) with data-augmentation optimized for QA Accuracy.

        \item \ourmodel~: our model described in \S \ref{sec:model-quartet} that optimizes for the best explanation structure.
                
    \squishend

\subsection{Explanation accuracy} \ourmodel~ is also the best model on explanation accuracy. Table \ref{tab:expl-results} shows the performance on $[i, j, d_i, d_e]$. \ourmodel~ also outperforms baselines on every component of the explanation. \ourmodel~ performs better at predicting $i$ than $j$. This trend correlates with human performance- picking on the second supporting sentence is harder because in a procedural text neighboring steps can have similar effects. 

\reviewed{We found that the explanation dataset does not contain substantial annotation artifacts for the \effectonly~ model to leverage (\effectonly~ $<$ \textsc{Majority})}

Table \ref{tab:many-examples} presents canonical examples of \ourmodel~ dev predictions.

\begin{table}[!h]
\centering
\resizebox{0.5\textwidth}{!}
{%
\begin{tabular}{|l|l|l|l|l|l|}
\hline
 & $acc_i$ & $acc_j$ & $acc_{d_i}$ & $acc_{d_e}$ & $acc_{expl}$ \\ \hline
\textsc{Random} & 12.50 & 12.50 & 33.33 & 33.33 & 22.91 \\ \hline
\effectonly~          & 32.77    & 32.77  & 33.50    & 44.82     & 36.00  \\\hline
\textsc{Majority} & 41.80 & 41.80 & 41.80 & 41.80 & 41.80 \\ \hline
\taggingmodel~ & 42.26 & 37.03 & 56.74 & 58.34 & 48.59 \\ \hline
\ourmodelplusnoexpl~   & 38.66  & 38.66  &   69.20    & 75.06   & 55.40  \\\hline
\ourmodel~ & \textbf{69.24} & \textbf{65.97} & \textbf{75.92} & \textbf{82.07} & \textbf{73.30} \\ \hline
\textsc{Human} & 75.90 & 66.10 & 88.20 & 96.30 & 81.63 \\ \hline
\end{tabular}%
}
\caption{Accuracy of the explanation structure ($i,j,d_i,d_e$). Overall explanation accuracy is $acc_{expl}$.
(Note that~\wiqabert{} and~\DataAug{} do not produce explanations).}
\label{tab:expl-results}
\end{table}

We also tried a simple bag of words and embedding vector based alignment between $q_p$ and $x_i$ in order to pick the most similar $x_i$. These baselines perform worse than random, showing that aligning $q_p$ and $x_i$ involves commonsense reasoning that the these models cannot address.

%% file: sections/downstream_qa.tex
\section{Downstream Task}
\label{sec:downstream_qa}

In this section, we investigate whether a good explanation structure leads to better end-task performance. \ourmodel~ advocates explanations as a first class citizen from which an answer can be derived. 

\subsection{Accuracy on a QA task} 
We compare against the existing SOTA on WIQA no-explanation task. Table \ref{tab:qa-accuracy} shows that \ourmodel~ improves over the previous SOTA \wiqabert~ by 7\%, achieving a new SOTA results. Both these models are trained on the same dataset\footnote{We used  the same code and parameters as provided by the authors of WIQA-BERT. The WIQA with-explanations dataset has about 20\% fewer examples than WIQA without-explanations dataset [http://data.allenai.org/wiqa/] This is because the authors removed about 20\% instances with incorrect explanations (e.g., where turkers didn’t have an agreement). So we trained both QUARTET and WIQA-BERT on exactly the same vetted dataset. This helped to increase the score of WIQA-BERT by 1.5 points.}. The major difference between \wiqabert~ and \ourmodel~ is that \wiqabert~ solves only the QA task, whereas \ourmodel~ solves explanations, and the answer to the QA task is derived from the explanation. 
Multi-tasking (i.e., explaining the answer) provides the gains to \ourmodel~. 

\begin{table}[!h]
\centering
{%
\begin{tabular}{|l|l|}
\hline
     & QA accuracy \\ \hline
    \textsc{Random} & 33.33 \\ \hline
    \textsc{Majority} & 41.80 \\ \hline
    \effectonly~          & 44.82\\\hline
    \taggingmodel~ & 58.34 \\ \hline
    \wiqabert~ & 75.19 \\ \hline
    \ourmodelplusnoexpl~  & 75.06 \\\hline
    \DataAug~ & 78.50 \\ \hline
    \ourmodel~ & \textbf{82.07} \\ \hline
    \textsc{Human} & 96.30 \\ \hline
\end{tabular}%
}
\caption{\ourmodel~ improves accuracy on the QA (end task) by 7\% points. }
\label{tab:qa-accuracy}
\end{table}

All the models get strong improvements over \textsc{Random} and \textsc{Majority}. The least performing model is \taggingmodel~. The space of possible sequences of correct labels is large, and we believe that the current training data is sparse, so a larger training data might help.  \ourmodel~ avoids this sparsity problem because rather than a sequence it learns on four separate explanation components.

Table \ref{tab:qa-accuracy-categories} presents the accuracy based on question types. \ourmodel~ achieves large gains over \wiqabert~ on the most challenging out-of-para questions. This suggests that \ourmodel~ improves the alignment of $q_p$ and $x_i$ that involves some commonsense reasoning.

\begin{table}[!h]
\small
\begin{tabular}{|l|l|l|l|l|}
\hline
Model             & in-para & out-of & no-effect & overall \\
             &  & para & &  \\\hline
\textsc{Random}    & 33.33     & 33.33      & 33.33                  & 33.33       \\\hline
\textsc{Majority} & 00.00     & 00.00      & 100.0                  & 41.80       \\\hline
\effectonly~          & 20.38     & 20.85      & 78.41   & 44.82   \\\hline
\wiqabert~              & 71.40     & 53.56      & 90.04                  & 75.19       \\\hline
\ourmodelplusnoexpl~   & 72.83   & 58.54          & 92.03   & 75.06       \\\hline
\ourmodel~          & \textbf{73.49}    & \textbf{65.65}      & \textbf{95.30}                  & \textbf{82.07}       \\\hline
\end{tabular}
\caption{\ourmodel~ improves accuracy over SOTA \wiqabert~ across question types.}
\label{tab:qa-accuracy-categories}
\end{table}

\subsection{Correlation between QA and Explanation} \ourmodel~ not only improves QA accuracy but also the explanation accuracy. We find that QA accuracy ($acc_{d_e}$ in Table \ref{tab:expl-results}) is positively correlated (Pearson coeff. 0.98) with explanation accuracy ($acc_{expl}$). This shows that if a model is optimized for explanations, it leads to better performance on end-task. Thus, with this result we establish that (at least on our task) models can make better predictions when forced to generate a sensible \textit{explanation structure}. An educational psychology study \cite{Dunlosky2013EducationPsychologyImprovingSL} hypothesizes that student performance improves when they are asked to explain while learning. However, their hypothesis is not conclusively validated due to lack of evidence. Results in Table \ref{tab:expl-results} hint that, at least on our task, machines that learn to explain, ace the end task.

%% file: sections/error_analysis.tex
\section{Error analysis}
\label{sec:analysis}
We analyze our model's errors (marked in \red{red}) over the dev set, and observe the following phenomena. 

\paragraph\noindent\textbf{1. Multiple explanations:} 
As mentioned in Section \ref{sec:problemdef}, more than one explanations can be correct. 
22\% of the incorrect explanations were reasonable, suggesting that overall explanation accuracy scores might under-estimate the explanation quality. The following example illustrates that while \texttt{gathering firewood} is appropriate when \texttt{fire is needed for survival}, one can argue that \texttt{going to wilderness} is less precise but possibly correct. 
\\
    
    \noindent\fbox{
    \small
    \parbox{0.46\textwidth}
    {
      \reallysquishlist
        \item[Gold:] need fire for survival $\rightarrow$ \blue{\texttt{(MORE/+)}} \bluebox{gather firewood} $\rightarrow$ \blue{\texttt{(MORE/+)}} \bluebox{build fire for warmth} $\rightarrow$ \blue{\texttt{(MORE/+)}} extensive camping trip
        
        \item[Pred:] need fire for survival $\rightarrow$ \blue{\texttt{(MORE/+)}} \redbox{go to wilderness} $\rightarrow$ \blue{\texttt{(MORE/+)}} \bluebox{build fire for warmth} $\rightarrow$ \blue{\texttt{(MORE/+)}} extensive camping trip
      \squishend
    }}

\begin{figure}[!ht]
        {\includegraphics[width=\columnwidth]{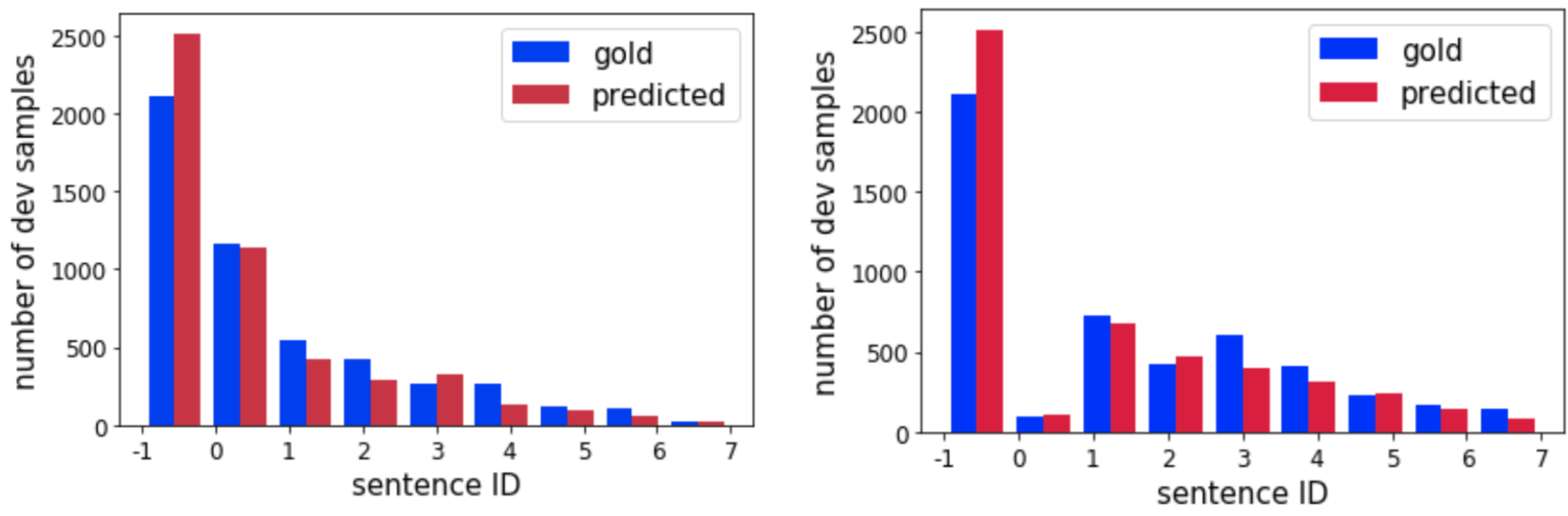}}
        \caption{Gold vs. predicted distribution of $i$ \& $j$ resp.}
        \label{fig:histogram-ij}
    \end{figure}

\paragraph \noindent \textbf{2. $i$, $j$ errors:}
Fig. \ref{fig:histogram-ij} shows that predicted and gold distributions of $i$ and $j$ are similar. Here, sentence id $= -1$ indicates no effect. The model has learned from the data to never predict $j < i$ without any hard constraints.

The model is generally good at predicting $i, j$ and in many cases when the model errs, the explanation seems plausible. Perhaps for the same underlying reason, human upper bound is not high on $i$ (75.9\%) and on $j$ (66.1\%). We show an example where $i$, $j$ are incorrectly predicted (in red), but sound plausible.

    \noindent\fbox{
    \small
    \parbox{0.46\textwidth}
    {
      \reallysquishlist
        \item[Gold:] ear is not clogged by infection $\rightarrow$ \blue{\texttt{(OPP/-)}} \bluebox{sound hits ear} $\rightarrow$ \blue{\texttt{(OPP/-)}} \bluebox{electrical impulse reaches brain} $\rightarrow$ \blue{\texttt{(OPP/-)}} more sound detected
        
        \item[Pred:] ear is not clogged by infection $\rightarrow$ \blue{\texttt{(OPP/-)}} \bluebox{sound hits ear} $\rightarrow$ \blue{\texttt{(OPP/-)}} \redbox{drum converts sound to electrical impulse} $\rightarrow$ \blue{\texttt{(OPP/-)}} more sound detected
      \squishend
    }}\\

\paragraph\noindent\textbf{3. $d_i$, $d_e$ errors:} 
When the model incorrectly predicts $d_i$, a major source of error is when `\opp~' is misclassified. ~70\% of the `\opp~' mistakes, should have been classified as `\cor~'. A similar trend is observed for $d_e$ but the misclassification of `\opp' is less skewed. Table \ref{tab:confusion-matrix-dide-overall} shows the confusion matrix of $d_i$ and of $d_e$ $in$ \{ \cor, \opp, \no~ \} . 
    \begin{table}[!h]
    \small
    \begin{tabular}{llll}
    \hline
              &  \no~  & \cor~  & \opp~         \\ \hline
    \no~        & 1972   &       91     &         47 \\
    \cor~          &  295   &      883     &        358 \\
    \opp~        & 226     &    492        &     639     \\
    \hline
    \end{tabular}
    \quad
    \begin{tabular}{llll}
    \hline
             &  \no~  & \cor~  & \opp~         \\ \hline
    \no~          & 1972   &       89     &         49 \\
    \cor~          &  261   &      909     &        295 \\
    \opp~       & 252     &    346        &     830     \\
    \hline
    \end{tabular}
\caption{Confusion matrix for $d_i$ (left) and $d_e$ overall (right). (gold is on x-axis, predicted on y-axis.)}
\label{tab:confusion-matrix-dide-overall}
\end{table}

    The following example shows an instance where `\opp~' is misclassified as `\pos~'. It implies that there is more scope for improvement here.\\
    
    \noindent\fbox{
    \small
    \parbox{0.44\textwidth}
    {
      \reallysquishlist
        \item[Gold:] less seeds fall to the ground  $\rightarrow$ \blue{\texttt{(OPP/-)}} \bluebox{seed falls to the ground} $\rightarrow$ \blue{\texttt{(OPP/-)}} \bluebox{seeds germinate} $\rightarrow$ \blue{\texttt{(MORE/+)}} fewer plants
        
        \item[Pred:] less seeds fall to the ground  $\rightarrow$ \blue{\texttt{(OPP/-)}} \bluebox{seed falls to the ground} $\rightarrow$ \blue{\texttt{(OPP/-)}} \bluebox{seeds germinate} $\rightarrow$ \red{\texttt{(OPP/-)}} fewer plants
        
      \squishend
    }}

\paragraph\noindent\textbf{4. in-para vs. out-of-para:} 
The model performs better on in-para questions (typically, linguistic variations) than out-of-para questions (typically, commonsense reasoning). Also see empirical evidence of this in Table \ref{tab:qa-accuracy-categories}.

The model is challenged by questions involving commonsense reasoning, especially to connect $q_p$ with $x_i$ in out-of-para questions. For example, in the following passage, the model incorrectly predicts \no~ (no effect) because it fails to draw a connection between \texttt{sleep} and \texttt{noise}:\\
    
    \noindent\fbox{
    \small
    \parbox{0.46\textwidth}
    {
        Pack up your camping gear, food.
        Drive to your campsite.
        Set up your tent.
        Start a fire in the fire pit.
        Cook your food in the fire.
        Put the fire out when you are finished.
        Go to sleep.
        Wake up ... 
        \vspace{2mm}
        \begin{des}
        \item[{\bf $q_p$:}] less noise from outside
        \item[{\bf $q_e$:}] you will have more energy
        \end{des}
    }} 
        \vspace{1mm}

Analogous to $i$ and $j$, the model also makes more errors between labels `\cor~' and  `\opp~' in out-of-para questions compared to in-para questions (39.4\% vs 29.7\%) -- see Table \ref{tab:confusion-di-qwise}. 

    \begin{table}[!h]
    \small
    \begin{tabular}{llll}
    \hline
              &  \no~  & \cor~  & \opp~         \\ \hline
    \cor~         &    29 &         295    &          78 \\
    \opp~         & 49 &        130        &     259 \\
    \hline
    \end{tabular}
    \quad
    \begin{tabular}{llll}
    \hline
             &  \no~  & \cor~  & \opp~         \\ \hline
    \cor~      &      266   &      588       &      280 \\
    \opp~      &  177    &     362      &       380 \\
    \hline
    \end{tabular}
\caption{Confusion matrix $d_i$ for in-para \& out-of-para}
\label{tab:confusion-di-qwise}
\end{table}

\cite{wiqa} discuss that some in-para questions may involve commonsense reasoning similar to out-of-para questions. The following is an example of an in-para question where the model fails to predict $d_i$ correctly because it cannot find the connection between \texttt{protected ears} and \texttt{amount of sound entering}.\\

    \noindent\fbox{
    \small
    \parbox{0.46\textwidth}
    {
      \reallysquishlist
        \item[Gold:] ears less protected  $\rightarrow$ \blue{\texttt{(MORE/+)}} \bluebox{sound enters  ear} $\rightarrow$ \blue{\texttt{(MORE/+)}} \bluebox{sound hits ear drum} $\rightarrow$ \blue{\texttt{(MORE/+)}} more sound detected
        
        \item[Pred:] ears less protected  $\rightarrow$ \red{\texttt{(OPP/-)}} \bluebox{sound enters the ear} $\rightarrow$ \red{\texttt{(OPP/-)}} \bluebox{sound hits ear drum} $\rightarrow$ \blue{\texttt{(MORE/+)}} more sound detected
        
      \squishend
    }}

\paragraph\noindent\textbf{5. Injecting background knowledge:}
To study whether additional background knowledge can improve the model, we revisit the out-of-para question that the model failed on. The model fails to draw a connection between \texttt{sleep} and \texttt{noise}, leading to an incorrect (no effect) `\no~' prediction. 

By adding the following relevant background knowledge sentence to the paragraph ``\texttt{sleep requires quietness and less noise}'', the model was able to correctly change probability mass from $d_e = $ `\no~' to `\cor~'. This shows that providing commonsense through Web paragraphs and sentences is a useful direction. \\
    
    \noindent\fbox{
    \small
    \parbox{0.46\textwidth}
    {
        Pack up your camping gear, food ... 
        \textbf{\textit{Sleeping requires quietness and less noise.}}
        Go to sleep.
        Wake up ... 
        \vspace{2mm}
        \begin{des}
        \item[{\bf $q_p$:}] less noise from outside
        \item[{\bf $q_e$:}] you will have more energy
        \end{des}
    }}

\section{Assumptions and Generality}

\ourmodel~ makes two simplifying assumptions: (1) explanations are assembled from the provided sentences (question + context), rather than generated, and (2) explanations are chains of qualitative, causal influences, describing how an end-state is influenced by a perturbation. Although these (helpfully) bound this work, the scope of our solution is still quite general: Assumption (1) is a common approach in other work on multihop explanation (e.g., HotpotQA), where authoritative sentences support an answer. In our case, we are the first to apply the same idea to chains of influences. Assumption (2) bounds \ourmodel~ to explaining the effects of qualitative, causal influences. However, this still covers a large class of problems, given the
importance of causal and qualitative reasoning in AI. The WIQA dataset provides the first large-scale dataset that exemplifies this class: given a qualitative influence, assemble a causal chain of events leading to a qualitative outcome. Thus \ourmodel~ offers a general solution within this class, as well as a specific demonstration on a particular dataset.

%% file: sections/conclusion.tex
\section{Conclusion}
Explaining the effects of a perturbation is critical, and we have presented the first system that can do this reliably. \ourmodel~ not only predicts meaningful explanations, but also achieves a new state-of-the-art on the end-task itself, leading to an interesting finding that models can make better predictions when forced to explain. \reviewed{Our work opens up new directions for future research: 1) Can additional background context from the Web improve explainable reasoning? 2) Can such structured explanations be applied to other NLP tasks? 
We look forward to future progress in this area.
}

%% file: sections/appendix.tex
\clearpage 
\section*{Appendix}

\subsection{Hyperparameter Tuning}
\label{subsec:appendix_hyperparam}
\ourmodel~ fine-tunes  \textsc{BERT}, allowing us to re-use the same hyperparameters as \textsc{BERT} with small adjustments in the recommended range \cite{Devlin2018BERT}. We use the \textsc{BERT}-base-uncased version with a hidden size of 768. We found the best hyperparameter settings by searching the space using the following hyperparameters. 
\begin{enumerate}
    \item weight decay = \{ 0.1, 0.01, 0.05 \}
    \item dropout = \{0.1, 0.2, 0.3 \}
    \item learning rate = \{1e-05, 2e-05, 5e-05\}
\end{enumerate}